\documentclass[10pt,peerreview,draftclsnofoot,onecolumn,a4paper]{IEEEtran}

\usepackage{amsmath} 
\usepackage{charter}
\usepackage{microtype}
\usepackage{algorithm,algpseudocode}
\usepackage{epsfig}
\usepackage{graphicx}
\usepackage[caption=false,font=footnotesize]{subfig}

\def\P{\mathbf{P}}

\ifCLASSINFOpdf
\else
\fi

\begin{document}

\title{Non-Local Patch Regression: \\ Robust Image Denoising in Patch Space}

\author{Kunal N. Chaudhury$^{*}$ \thanks{$^{*}$Program in Applied and Computational Mathematics (PACM), Princeton University, Princeton, NJ 08544, USA (kchaudhu@math.princeton.edu). 
K. Chaudhury was supported by the Swiss National Science Foundation under grant PBELP2-$135867$.} \hspace{6mm} 
Amit Singer$^{**}$\thanks{$^{**}$PACM and Department of Mathematics, Princeton University, Princeton, NJ 08544, USA (amits@math.princeton.edu).
A. Singer was partially supported by Award Number DMS-0914892 from the NSF, by Award Number FA9550-09-1-0551 from AFOSR, by Award Number R01GM090200 from the National Institute of General Medical Sciences, 
and by the Alfred P. Sloan Foundation.} }

\maketitle

\begin{abstract}
It was recently demonstrated in \cite{Chaudhury2012} that the denoising performance of Non-Local Means (NLM) can be improved
at large noise levels by replacing the mean by the robust Euclidean median. Numerical experiments on synthetic and natural images
showed that the latter consistently performed better than NLM beyond a certain noise level, and significantly so for images with sharp edges. The Euclidean mean and median can be put into a common
regression (on the patch space) framework, in which the $\ell_2$ norm of the residuals is considered in the former, while the $\ell_1$ norm is considered in the latter.
The natural question then is what happens if we consider $\ell_p \ (0<p<1)$ regression? We investigate this possibility
in this paper. 
\end{abstract}


\section{Introduction}

In the last few years, some very effective frameworks for image restoration have been proposed that exploit non-locality (long-distance correlations) in images, and/or use patches instead of pixels to robustly 
compare photometric similarities. The archetype algorithm in this regard is the Non-Local Means (NLM) \cite{BCM2005}. The success of NLM triggered a huge amount 
of research, leading to state-of-the-art algorithms that exploit non-locality and/or the patch model in specialized ways; e.g., see \cite{Kervrann2006,Giloba2008,BM3D,Aharon2008,Peyre2011,Chatterjee2012,Deledalle2012}, 
to name a few. We refer the interested reader to \cite{BCM2010,Chatterjee2012} for detailed reviews. Of these, the best performing method till date is perhaps the hybrid BM3D algorithm \cite{BM3D}, which effectively combines the NLM framework with other classical algorithms. 

To setup notations, we recall the working of NLM. Let $u = (u_i)$ be some linear indexing of the input noisy image. The standard setting is that $u$ is the corrupted
version of some clean image $f = (f_i)$,
\begin{equation}
\label{model}
u_i = f_i + \sigma z_i,
\end{equation}
where $(z_i)$ is iid $\mathcal{N}(0,1)$. The goal is to estimate (approximate) $f$ from the noisy measurement $u$, 
possibly given a good estimate of the noise floor $\sigma$. In NLM, the restored image $\hat{u} = (\hat{u}_i)$ is computed using the simple formula
\begin{equation}
\label{NLM_formula}
\hat{u}_i  = \frac{\sum_{j \in S(i)} w_{ij} u_j}{\sum_{j \in S(i)} w_{ij} },
\end{equation}
where $w_{ij}$ is some weight (affinity) assigned to pixels $i$ and $j$. Here $S(i)$ is the neighborhood of pixel $i$ over which the averaging is performed. 
To exploit non-local correlations, $S(i)$ is ideally set to the whole image domain. In practice, however, one restricts $S(i)$ to a geometric neighborhood, e.g., to a 
sufficiently large window of size $S \times S$ around $i$ \cite{BCM2005}. The other idea in NLM is to  set the weights using image patches centered around each pixel. In particular, for a given pixel $i$, 
let $\P_i$ denote the restriction of $u$ to a square window around $i$. Letting $k$ be the length of this window, this associates every pixel $i$ with a point $\P_i$ in $\mathbf{R}^{k^2}$ (the patch space). 
The weights in standard NLM are set to be
\begin{equation}
\label{Weights}
 w_{ij} = \exp\Big(- \frac{1}{h^2}\lVert \P_i - \P_j \rVert^2  \Big),
\end{equation}
where $\lVert \P_i - \P_j \rVert$ is the Euclidean distance between $\P_i$ and $\P_j$ as points in $\mathbf{R}^{k^2}$, and $h$ is a smoothing parameter. 
Along with non-locality, it is the use of patches that makes NLM more robust in comparison to pixel-based neighborhood filters \cite{Yaroslavsky1985,Smith1997,Tomasi1998}. 

Recently, it was demonstrated in \cite{Chaudhury2012} that the denoising performance of NLM can be improved (often substantially for images with sharp edges) by replacing the $\ell_2$ regression in NLM with the more robust $\ell_1$ regression. More precisely, given weights $w_{ij}$, note that \eqref{NLM_formula} is equivalent to performing the following regression (on the patch space):
\begin{equation}
\label{L2_regression}
\hat{\P}_i =  \arg \ \min_{\P } \  \sum_{j \in S(i)} w_{ij} \lVert \P - \P_j \rVert^2,
\end{equation}
and then setting $\hat{u}_i$ to be the center pixel in $\hat{\P}_i$. Indeed, this reduces to \eqref{NLM_formula} once we write the regression in terms of the center pixel $\hat{u}_i$. The
idea in \cite{Chaudhury2012} was to use $\ell_1$ regression instead, namely, to compute
\begin{equation}
\label{L1_regression}
\hat{\P}_i =  \arg \ \min_{\P } \  \sum_{j \in S(i)} w_{ij} \lVert \P - \P_j \rVert,
\end{equation}
and then set $\hat{u}_i$ to be the center pixel in $\hat{\P}_i$. Note that \eqref{L1_regression} is a convex optimization, and the minimizer (the Euclidean median) is unique when $k > 1$ \cite{Milasevic1987}. 
The resulting estimator was called the Non-Local Euclidean Medians (NLEM). A numerical scheme was proposed in \cite{Chaudhury2012} for computing the Euclidean median using a sequence of weighted least-squares. It was 
demonstrated that NLEM performed consistently better than NLM on a large class of synthetic and natural images, as soon as the noise was above a certain 
threshold. More specifically, it was shown that the bulk of the improvement in NLEM came from pixels situated close to edges. An inlier-outlier model of the patch space around an edge was proposed, and the improvement was 
attributed to the robustness of \eqref{L1_regression} in the presence of outliers \cite{HR2009}. 

In this paper, we show how a simple extension of the above idea can dramatically improve the denoising performance of NLM, and even that of NLEM. This is the content of Section
II. In particular, a general optimization and algorithmic framework is provided that includes NLM and NLEM as special cases. Some numerical results on synthetic and natural images are provided in Section III to justify our claims. 
Possible extensions of the present work are discussed in Section IV.

\section{Non-Local Patch Regression}

\subsection{Robust patch regression}

 It is well-known that $\ell_1$ minimization is more robust to outliers than $\ell_2$ minimization. A simple argument is that the \textit{unsquared} 
residuals $\lVert \P - \P_j \rVert$ in \eqref{L1_regression} are better guarded against the aberrant data points compared to the squared residuals $\lVert \P - \P_j \rVert^2$. The former 
tends to better suppress the large residuals that may result from outliers. This basic principle of robust statistics can be traced back to the works of von Neumann, Tukey \cite{Tukey1977}, and Huber \cite{HR2009}, and lies at 
the heart of several recent work on the design of robust estimators; e.g., see \cite{Tropp2012}, and the references therein. 

A natural question is what happens if we replace the $\ell_1$ regression in \eqref{L1_regression} by $\ell_{(p < 1)}$ regression? In general, one could consider the following class of problems:
\begin{equation}
\label{Lp_regression}
\hat{\P}_i =  \arg \ \min_{\P } \  \sum_{j \in S(i)} w_{ij} \lVert \P - \P_j \rVert^p.
\end{equation}
The intuitive idea here is that, by taking smaller values of $p$, we can better suppress the residuals $\lVert \P - \P_j \rVert$ induced by the outliers. 
This should make the regression even more robust to outliers, compared to what we get with $p=1$. We note that a flip side of setting $p < 1$ is that
\eqref{Lp_regression} will no longer be  convex (this is essentially because $t \mapsto |t|^p$ is convex if and only if $p \geq 1$), and it is in general difficult to find the global minimizer of a non-convex functional. However, 
we do have a good chance of finding the global optimum if we can
initialize the solver close to the global optimum. The purpose of this note is to numerically demonstrate that, for all sufficiently large $\sigma$, the $\hat{u}$ obtained by solving \eqref{Lp_regression} (and
letting $\hat{u_i}$ to be the center pixel in $\hat{\P}_i$) results in a more robust approximation of $f$ as $p \rightarrow 0$, than what is obtained using NLM. 
Henceforth, we will refer to \eqref{Lp_regression} as Non-Local Patch Regression (NLPR), where $p$ is  generally allowed to take values in the range $(0,2]$.

\subsection{Iterative solver}

The usefulness of the above idea actually stems from the fact that there exists a simple iterative solver for \eqref{Lp_regression}. In fact, the idea was 
influenced by the well-known connection between `sparsity' and `robustness', particularly the use of $l_{(p < 1)}$ minimization for best-basis selection and exact sparse recovery \cite{Rao1999,Chartrand2007,Saaba2010}. 
We were particularly motivated by the iteratively reweighted least squares (IRLS) approach of Daubechies et al. \cite{DDFG2009}, and a regularized version of IRLS developed by Chartrand for non-convex optimization \cite{Chartrand2007,Chartrand2008}. We will adapt the regularized IRLS algorithm in \cite{Chartrand2007,Chartrand2008} for solving \eqref{Lp_regression}. The exact working of this iterative solver is as follows. We use 
the NLM estimate to initialize the algorithm, that is, we set 
\begin{equation}
\label{init}
\P^{(0)} = \frac{\sum_{j \in S(i)} w_{ij} \P_j}{\sum_{j \in S(i)} w_{ij} }.
\end{equation}
Then, at every iteration $k \geq 1$, we write $\lVert \P - \P_j \rVert^p = \lVert \P - \P_j \rVert^2 \cdot \lVert \P - \P_j \rVert^{p-2}$ in \eqref{Lp_regression}, and use the current estimate to approximate this 
by $\lVert \P - \P_j \rVert^2 \cdot \lVert \P^{(k-1)} - \P_j \rVert^{p-2}$. This gives us the surrogate least-squares
problem
\begin{equation}
\label{surrogate}
\P^{(k)} =  \arg \ \min_{\P } \  \sum_{j \in S(i)} w_{ij} \frac{\lVert \P - \P_j\rVert^2}{ \left(\lVert \P^{(k-1)} - \P_j\rVert^2 + \varepsilon^{(k)} \right)^{1-p/2}}.
\end{equation}
Here $\varepsilon^{(k)} > 0$ is used as a guard against division by zero, and is gradually shrunk to zero as the iteration progresses. We refer the reader to \cite{Chartrand2007} for details. 
The solution of \eqref{surrogate} is explicitly given by
\begin{equation}
\label{linear}
\P^{(k)} =  \frac{ \sum_{j \in S(i)} w_{ij} \mu^{(k)}_j \P_j}{ \sum_{j \in S(i)} w_{ij} \mu^{(k)}_j},
\end{equation}
where 
$$\mu^{(k)}_j =  (\lVert \P^{(k-1)} - \P_j\rVert^2 + \varepsilon^{(k)} )^{p/2-1}.$$ 
The minimizer of \eqref{Lp_regression} is taken to be the limit of the iterates, assuming that it exists. While we cannot provide any guarantee on local convergence at this point, we note that
\eqref{linear} can be expressed as a gradient descent step (with appropriate step size) of a smooth surrogate of \eqref{Lp_regression}. This interpretation leads to the well-known 
Weiszfeld algorithm (for the special case $p=1$), which is known to converge linearly \cite{W1937,Hartley2011}. Alternatively, one could adapt more sophisticated IRLS algorithms (e.g., the one in \cite{DDFG2009}), which come with proven guarantees on local convergence, to the case $p<1$.

The overall computational complexity of 
NLPR is $O(k^2 S^2 I)$ per pixel, where $I$ is the average number of iterations. For NLM, the complexity is $O(k^2 S^2)$ per pixel. For a given convergence accuracy, we have noticed that $I$ increases as $p$ decreases. 
In particular, a large number of iterations are required in the non-convex regime $0 < p < 0.4$. In this case, we halt the computation after a sufficiently large number of iterations.

\begin{algorithm}
\caption{Non-Local Patch Regression (NLPR)}
\label{algo1}
\begin{algorithmic}
     \State \textbf{Input}: Noisy image $u = (u_i)$, and parameters $h, S, k, p$.
     \State \textbf{Return}: Denoised image $\hat{u} = (\hat{u}_i)$.
     \State (1) Extract patch $\P_i$ of size $k \times k$ at every pixel $i$. 
     \State (2) For every pixel $i$, do
       \State \hspace{1mm}   (a)  Set $w_{ij} = \exp(-\lVert \P_i - \P_j \rVert^2/h^2)$ for every $j \in S(i)$.
       \State \hspace{1mm}   (b)  Sort $w_{ij}, j \in S(i),$ in non-increasing order.
       \State \hspace{1mm}   (c)  Let $j_1,j_2,\ldots,j_{S^2}$ be the re-indexing of $j \in S(i)$ as per the above order. 
       \State \hspace{1mm}   (d)  Find patch $\P$ that minimizes $\sum_{t = 1}^{[S^2/2]} w_{ij_{t}} \lVert \P - \P_j \rVert^p$.
       \State \hspace{1mm}   (e)  Set $\hat{u}_i$ to be  the center pixel in $\P$.
\end{algorithmic}
\end{algorithm}	

\subsection{Robustness using $k$-nearest neighbors}

We noticed in \cite{Chaudhury2012} that a simple heuristic often provides a remarkable improvement in the performance of NLM.
In \eqref{NLM_formula}, one considers all patches $\P_j, j \in S(i),$ drawn from the geometric neighborhood of pixel $i$. 
However, notice that when a patch is close to an edge, then roughly half of its neighboring patches are on one side (the correct side) of the edge. 
Following this observation, we consider only the top $50\%$ of the the neighboring patches that have the largest weights. That is, the selected patches correspond to the $[r/2]$-nearest neighbors of $\P_i$ in the patch space, 
where $r=|S(i)|$. While this tends to inhibit the diffusion at low noise levels (in smooth regions), it was demonstrated in \cite{Chaudhury2012} that it can significantly improve the robustness 
of NLM and NLEM at large $\sigma$. We will also use this heuristic in NLPR. The overall scheme is summarized in Algorithm \ref{algo1}. We use $S(i)$ to denote a window of size $S \times S$ centered at pixel $i$
in the algorithm.

\section{Numerical Experiments}

To understand the denoising performance of NLPR, we provide some limited results on synthetic and natural images. The main theme of our investigation would be to understand how
the performance of NLPR changes with the regression index $p$. For a quantitative comparison of the denoising results, we will use the standard peak-signal-to-noise ratio (PSNR).
For an $N$-pixel image, with intensity scaled to $[0,1]$, this is defined to be $-10\log_{10}(\epsilon)$, where $\epsilon = (1/N) \sum_{i=1}^N (\hat{u}_i - f_i)^2$.

\begin{figure}[!htp]
  \centering
   \subfloat{\includegraphics[width=0.3\linewidth]{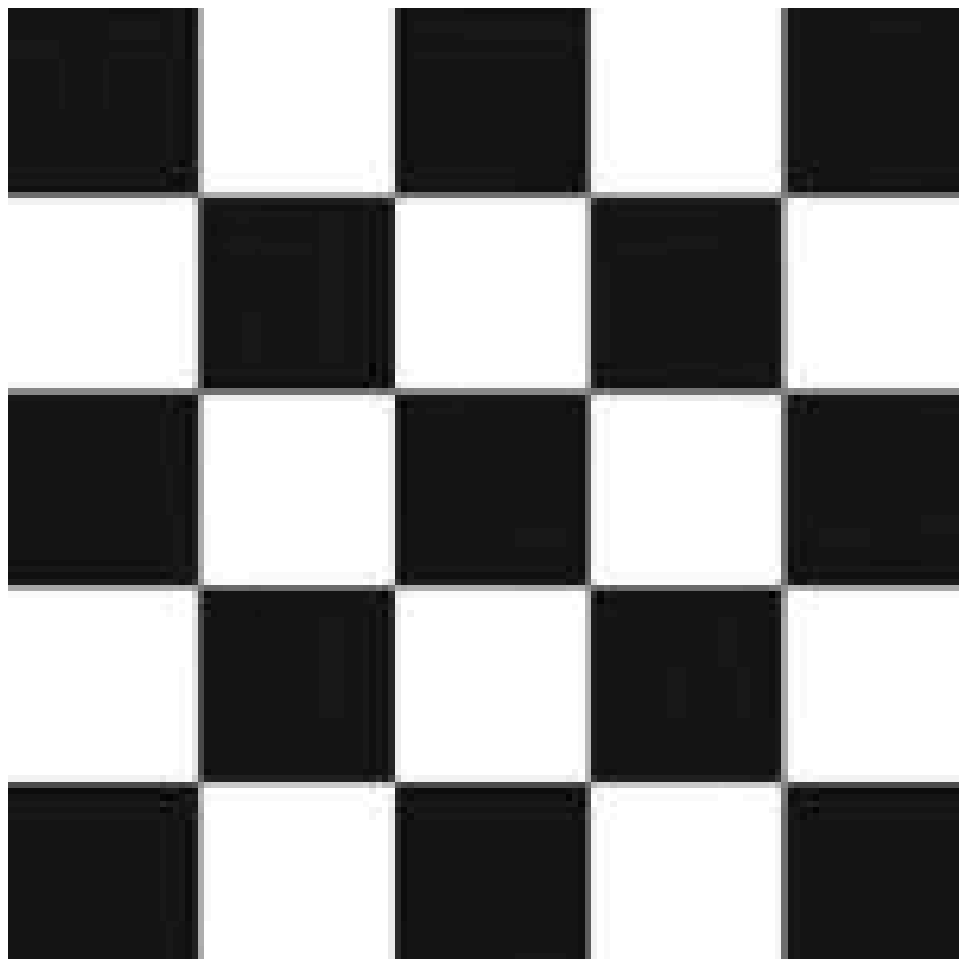}} 
  \subfloat{\includegraphics[width=0.5\linewidth]{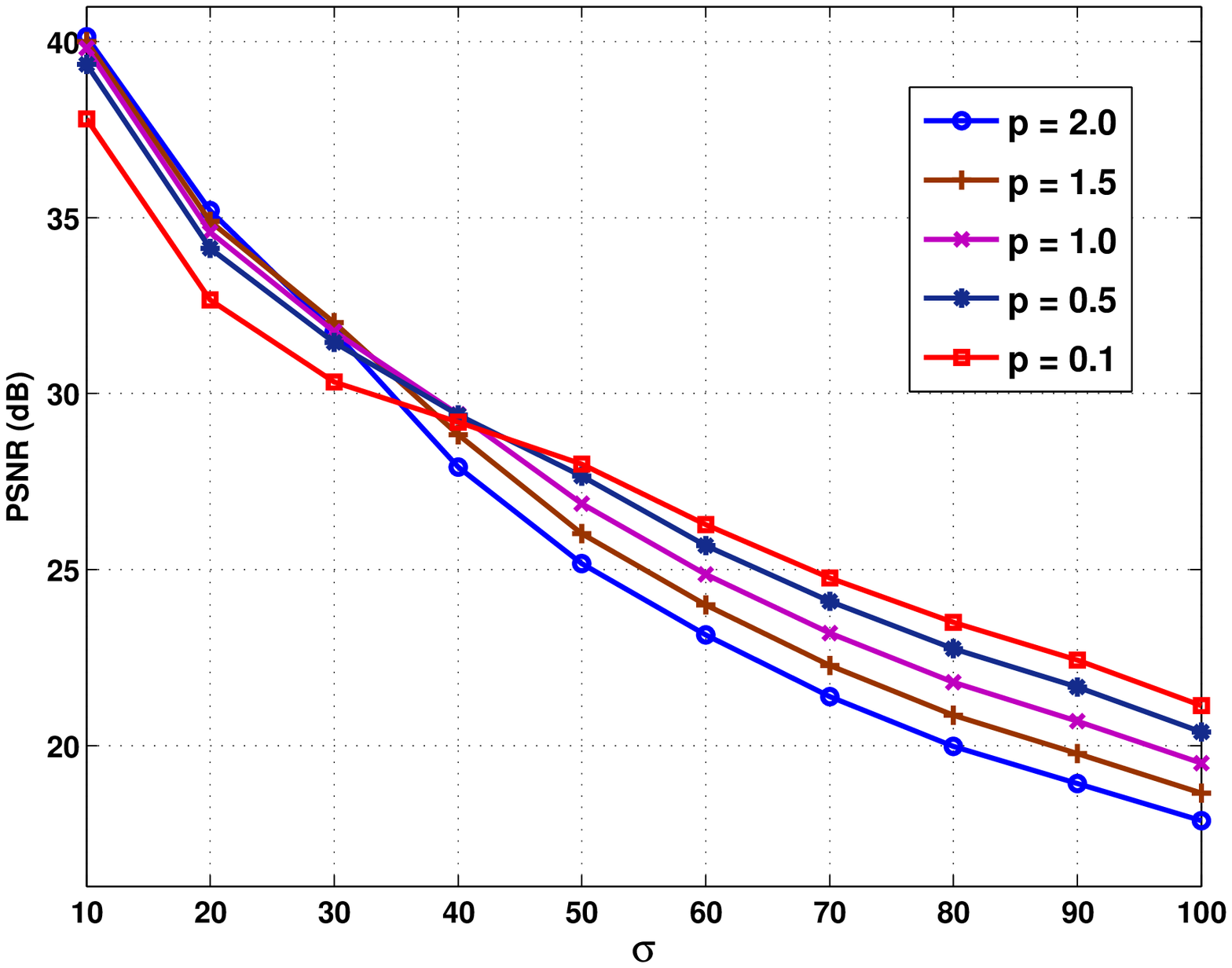}} 
  \caption{Left: Test image \textit{Checker}. Right: PSNRs obtained using NLPR for the test image \textit{Checker} at different $\sigma$ and $p$. To compare the different regressions, 
  we skipped steps (b) and (c) in Algorithm \ref{algo1}, i.e., we consider all the neighboring patches, and not just the top $50\%$.}
  \label{PSNR_sigma_p}
\end{figure}

We first consider the test image of \textit{Checker} used in \cite{Chaudhury2012}. This serves as a good model for simultaneously testing the denoising quality in smooth regions and in the vicinity of edges. 
We corrupt \textit{Checker} as per the noise model in \eqref{model}. We then compute the denoised image using Algorithm \ref{algo1}, with the exception
that we skip steps (b) and (c), that is, we use the full neighborhood $S(i)$. We initialize the iterations of the IRLS solver using \eqref{init}. 
For all the experiments in this paper, we fix the parameters to be $S = 21, k = 7,$ and $h = 10\sigma$. These are the settings originally proposed in \cite{BCM2005}. The results obtained using these 
settings are not necessarily optimal, and other settings could have been used as well. The point is to fix all the parameters in Algorithm \ref{algo1}, except $p$. This means 
that the same $w_{ij}$ are used for different $p$. We now run the above denoising experiment for $\sigma = 10, 20, \ldots, 100$, and for $p = 0.1, 0.5, 1, 1.5, 2$. 

\begin{figure}[!htp]
  \centering
  \subfloat[Clean and noisy edge ($\sigma = 0.3$).]{\label{em1}\includegraphics[width=0.5\linewidth]{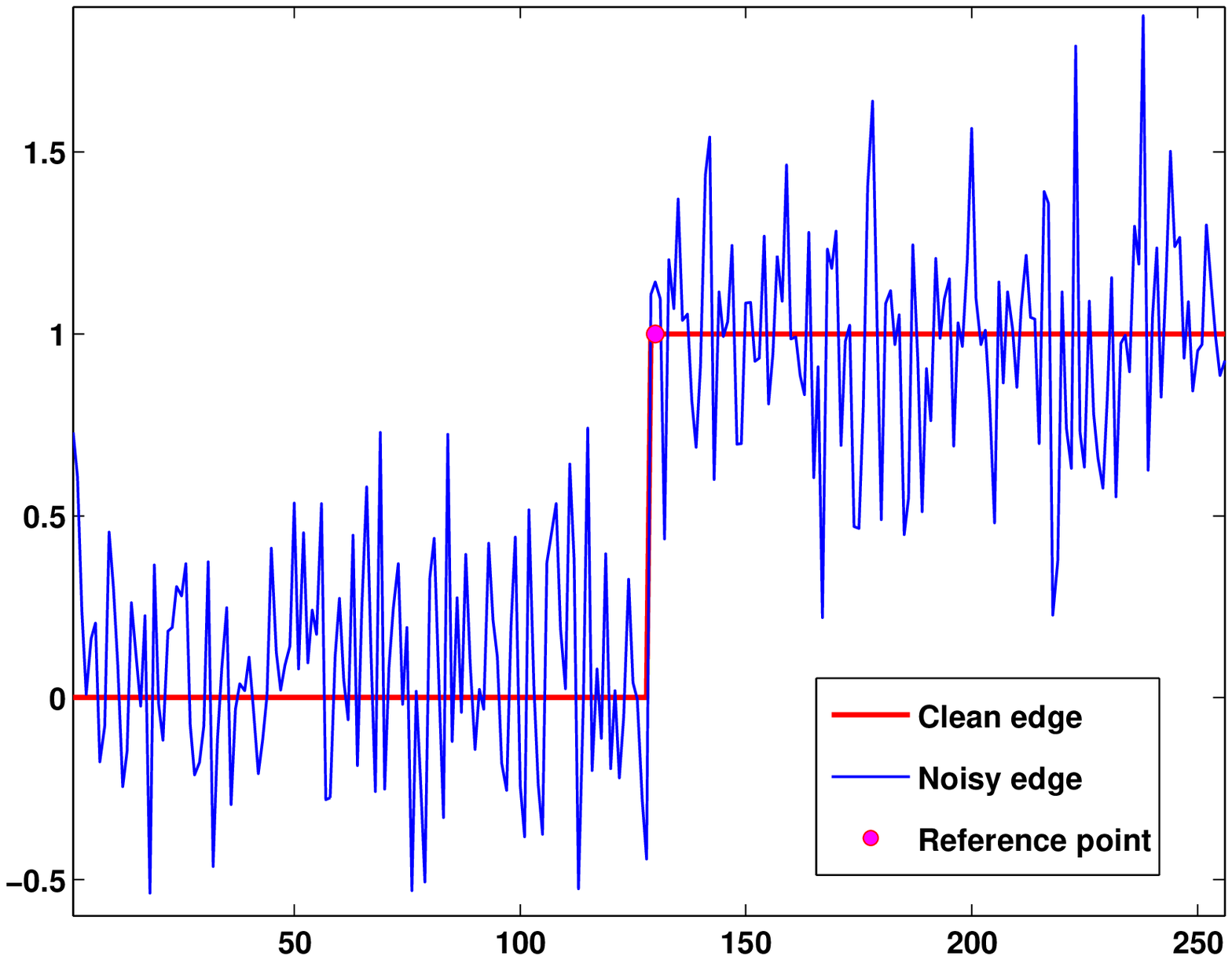}} 
  \subfloat[Weights.]{\label{em2}\includegraphics[width=0.5\linewidth]{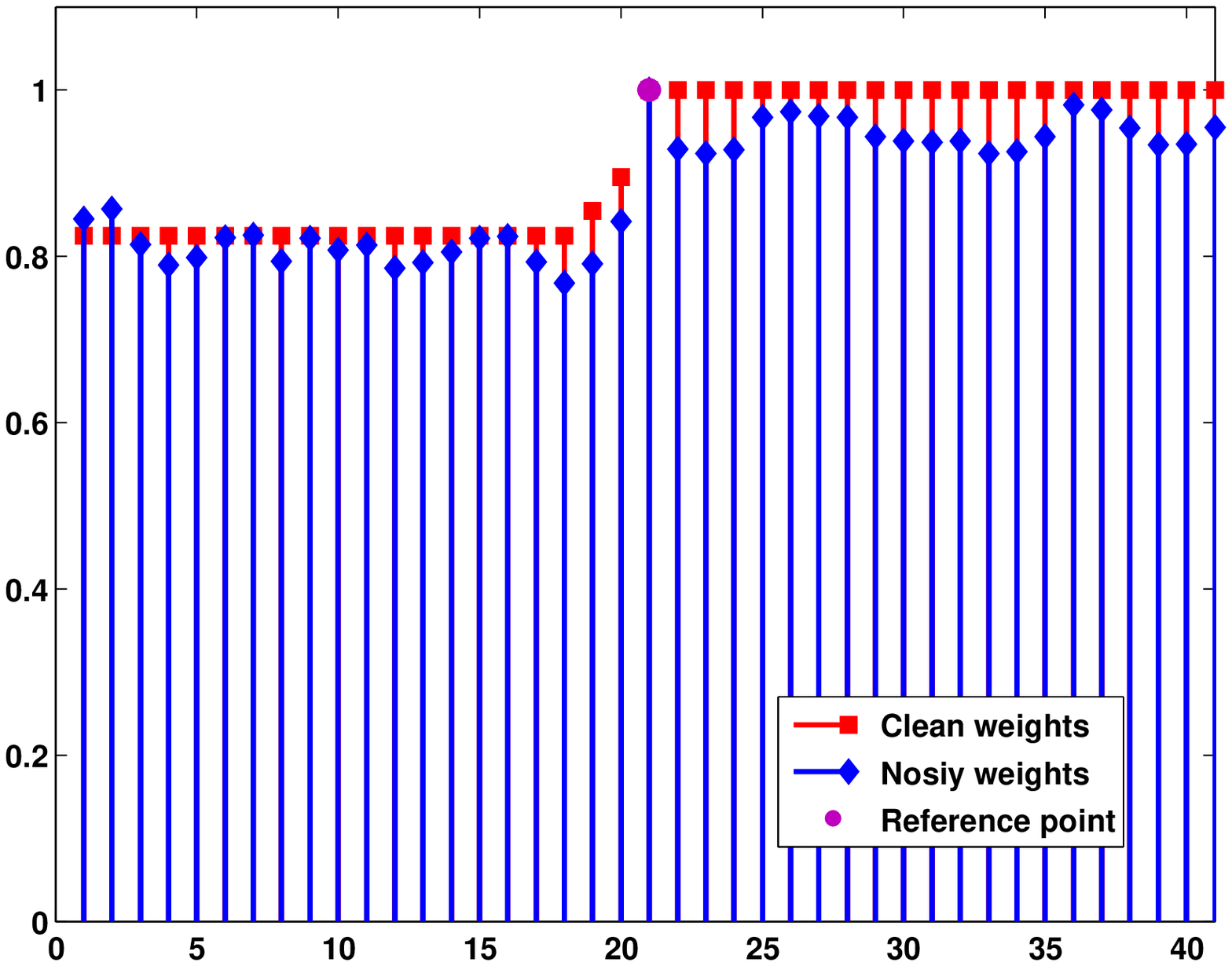}}
  \caption{Ideal edge of length $256$ used to evaluate the performance of NLPR. Each patch $\P_m$ is formed using $3$ points around position $m$, i.e., the patch space is of dimension $3$ (shown in Figure \ref{In-Out}). 
  The reference patch $\P_i$ corresponds to the position $i=130$ (situated close to the edge). The weights in \ref{em2} are computed between the reference patch and the neighboring patches $\P_j, j \in [i - 20, i +20]$.}
    \label{edgeModel}
\end{figure}

The results are shown in Figure \ref{PSNR_sigma_p}. We notice that, beyond a certain noise level, NLPR performs better when $p$ is close to zero. In fact, the PSNR increases 
gradually from $p=2$ to $p=0.1$, for a fixed $\sigma$. At lower noise levels, the situation reverses completely, and NLPR tends to perform better around $p=2$. 
A possible explanation is that the true neighbors in patch space are well identified at low noise levels, and since the noise is Gaussian, $\ell_2$ regression gives statistically optimal results.


\begin{figure}[!htp]
  \centering
  \subfloat[3d patch space.]{\label{io1}\includegraphics[width=0.5\linewidth]{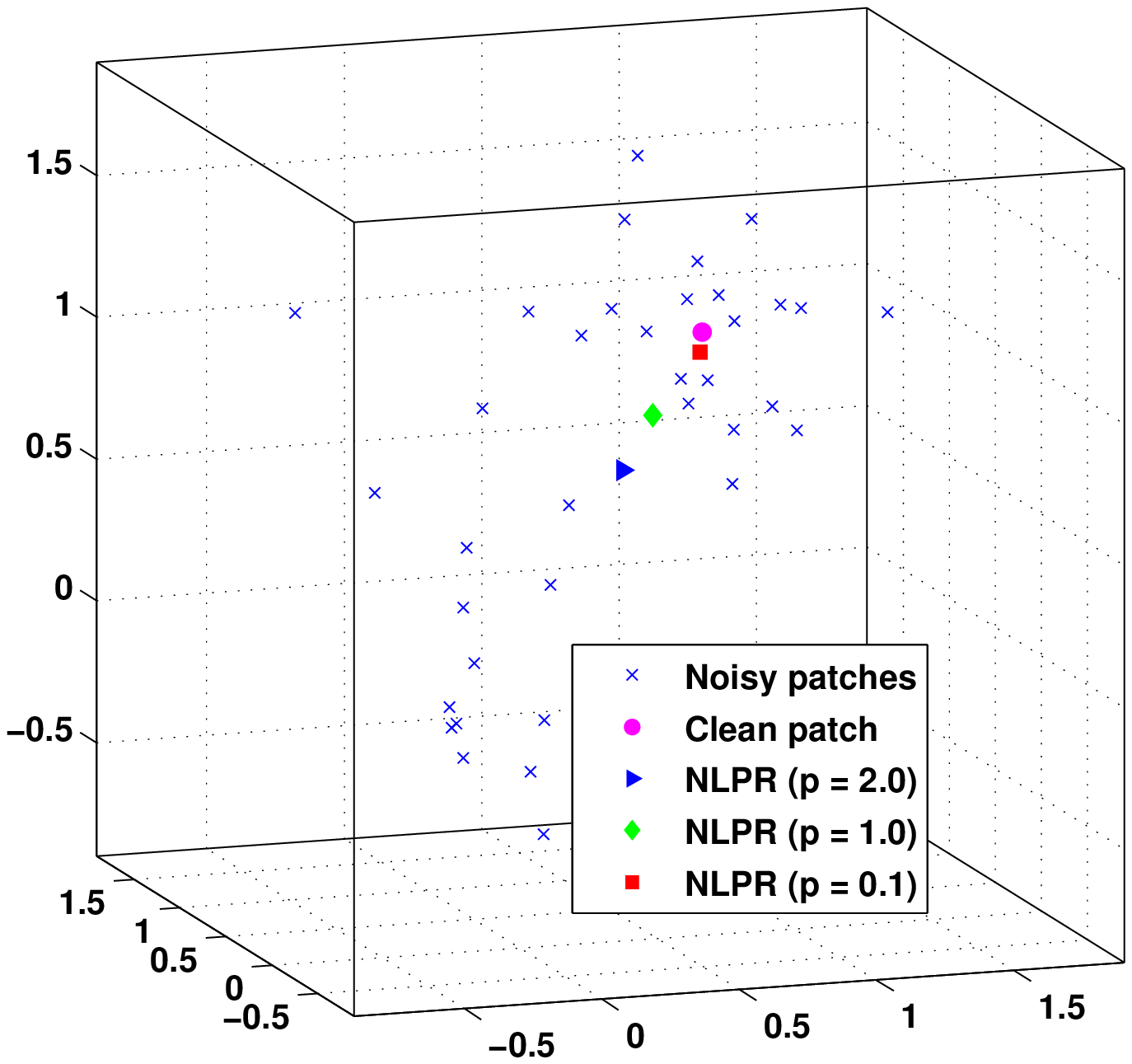}}  
  \subfloat[First $2$ coordinates.]{\label{io2}\includegraphics[width=0.5\linewidth]{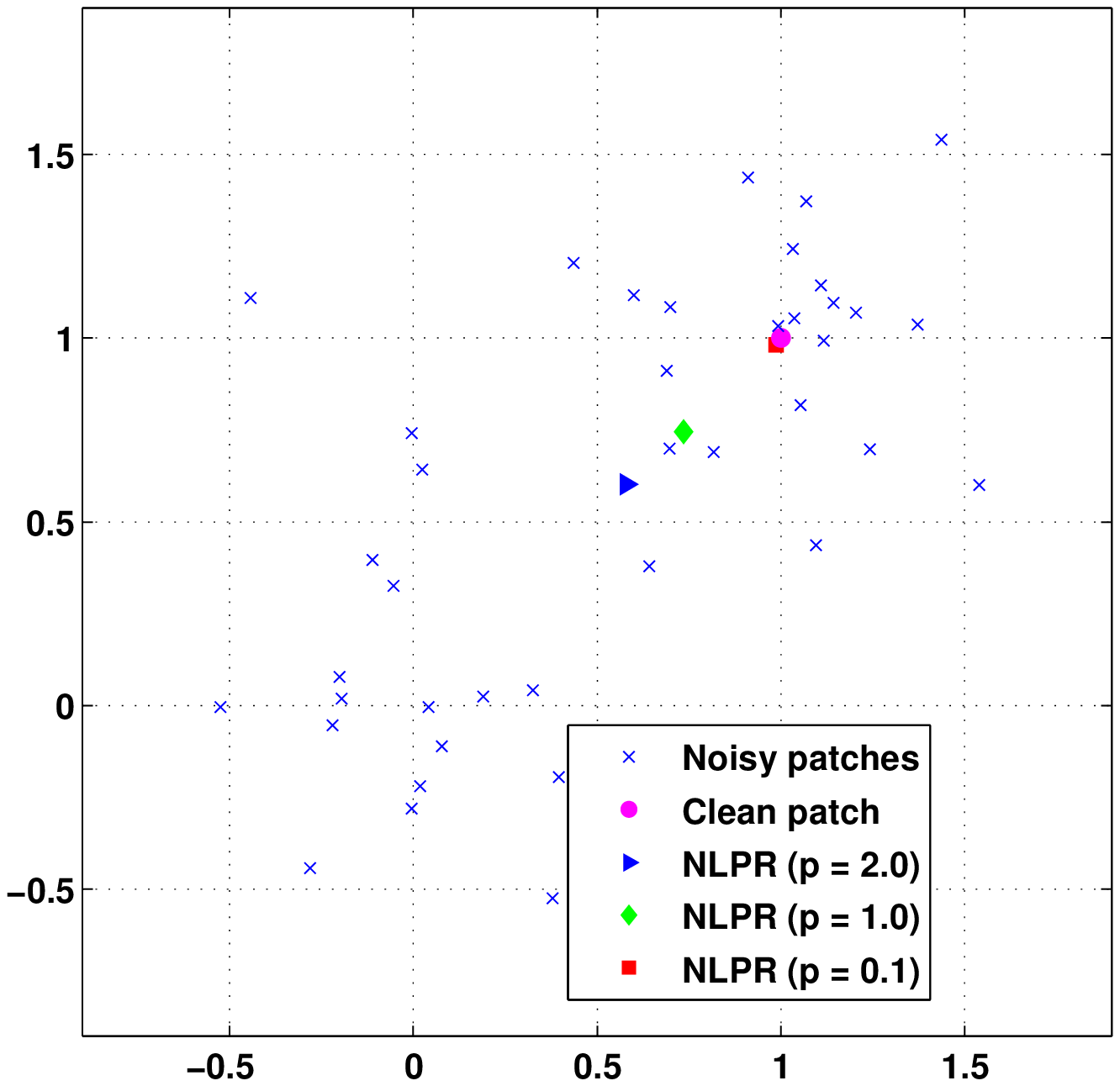}}
  \caption{Inlier-outlier model of the patch space for the reference point marked in Figure \ref{edgeModel}. Note that the estimate returned by NLPR gets better as $p$ goes from $2$ to $0.1$.
  This is consistent with the results in Figure \ref{PSNR_sigma_p}.}
  \label{In-Out}
\end{figure} 

An analysis of the above results shows us that, as $p \rightarrow 0$, the bulk of the improvement comes from pixels situated in the vicinity of edges. A similar observation was also made
in \cite{Chaudhury2012} for NLEM. To understand this better, we recall the ideal $0$-$1$ edge model used in \cite{Chaudhury2012}. This is shown in Figure \ref{em1}. We add noise of strength $\sigma=0.3$ to the edge,
and denoise it using NLPR. We examine the regression at a reference point situated just right to the edge (cf. Figure \ref{em2}). The patch space at this point
is specified using $k=3$ and $S=41$. The distribution of patches is shown in Figure \ref{In-Out}. Note that the patches are clustered around the centers $A=(0,0,0)$ and $B=(1,1,1)$. 
For the reference point, the points around $A$ are the outliers, while the ones around $B$ are the inliers. We now perform $\ell_p$ regression on this distribution for $p=0.1,1,$ and $2$. The results obtained (Algorithm \ref{algo1}, steps (b) and (c) skipped) from a single noise realization are shown in Figure \ref{In-Out}. The exact values of the estimate in this case are $0.61$ ($p=2$), $0.75$ ($p=1$), and $0.98$ ($0.1$). 
The average estimate over $10$ noise realizations are $0.58$ ($p=2$), $0.82$ ($p=1$), and $0.95$ ($p=0.1$). 

\begin{figure}[!htp]
  \centering
  \subfloat{\includegraphics[width=0.6\linewidth]{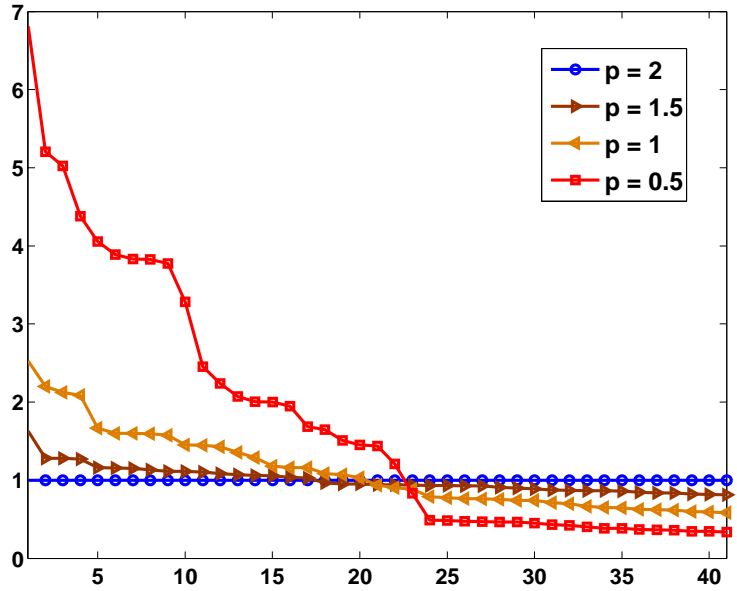}} 
  \caption{The multipliers $\mu_j^{(\infty)}, j \in S(i),$ in \eqref{linear} (sorted in non-increasing order) for the experiment with the ideal edge.}
  \label{mu}
\end{figure}

We note that the working of the IRLS algorithm provides some insight into the robustness of $\ell_p$ regression. Note that when $p=2$ (NLM), the reconstruction in \eqref{Lp_regression} is linear; the contribution of each 
noisy patch $\P_j$ is controlled by the corresponding weight $w_{ij}$. On the other hand, the reconstruction is non-linear when $p < 2$. The contribution of each $\P_j$ is 
controlled not only by the respective weights, but also by the multipliers $\mu^{(k)}_j$. 
In particular, the limiting value of the multipliers dictate the contribution of each $\P_j$ in the final reconstruction. Figure \eqref{mu} gives the distribution of the sorted multipliers (at convergence) for the experiment 
described above. In this case, the large multipliers correspond to the inliers, and the small multipliers correspond to the outliers. Notice that when $p=0.5$, the tail part of the multipliers (outliers) has much smaller 
values (close to zero) compared to the leading part (inliers). In a sense, the iterative algorithm gradually `learns' the outliers from the patch distribution as the iteration progresses, which are finally 
taken out of estimation.

\begin{figure}[!htp]
  \centering
  \subfloat[\textit{Barbara}.]{\includegraphics[width=0.42\linewidth]{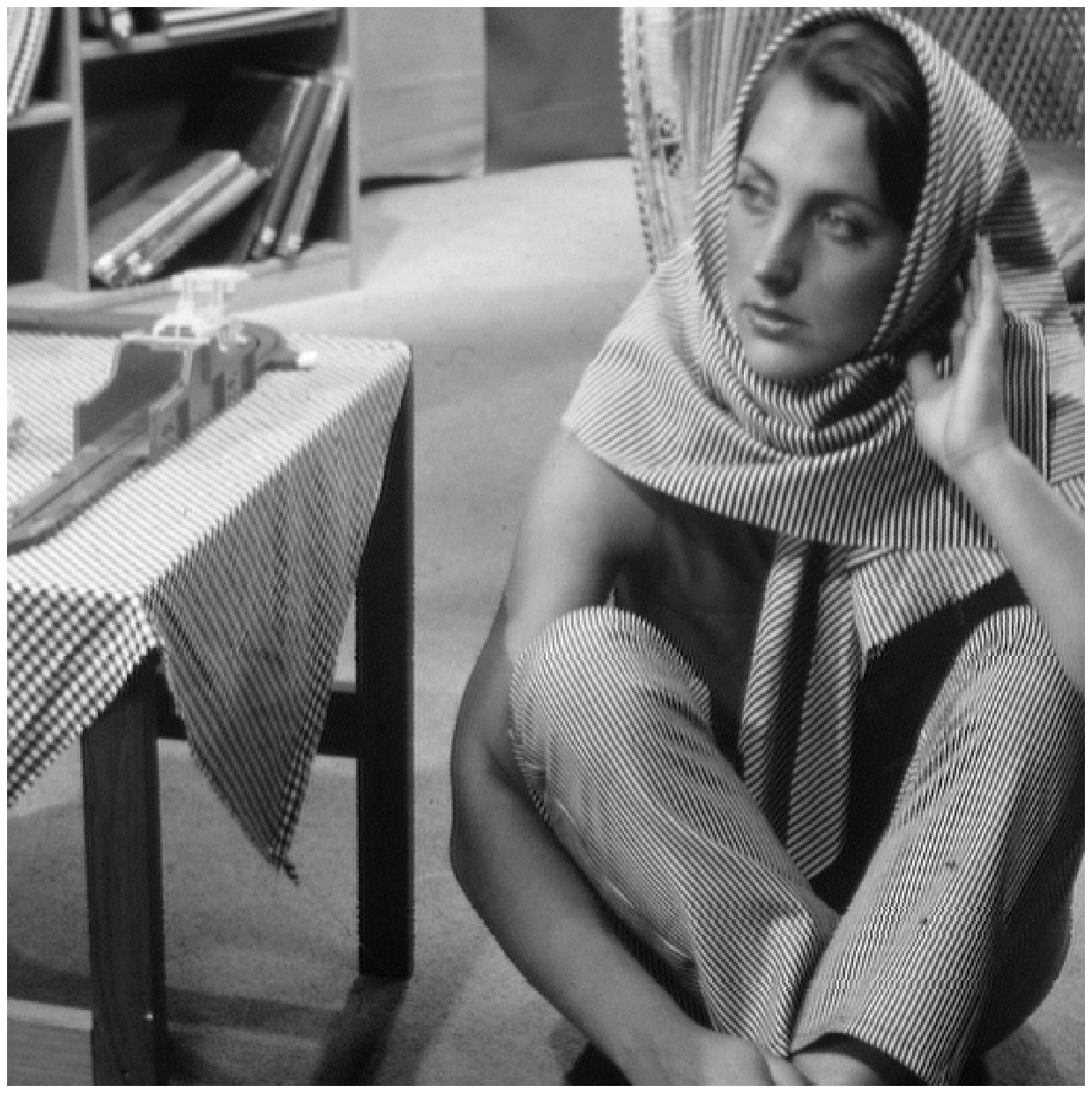}} 
  \subfloat[Corrupted ($\sigma =40$).]{\includegraphics[width=0.42\linewidth]{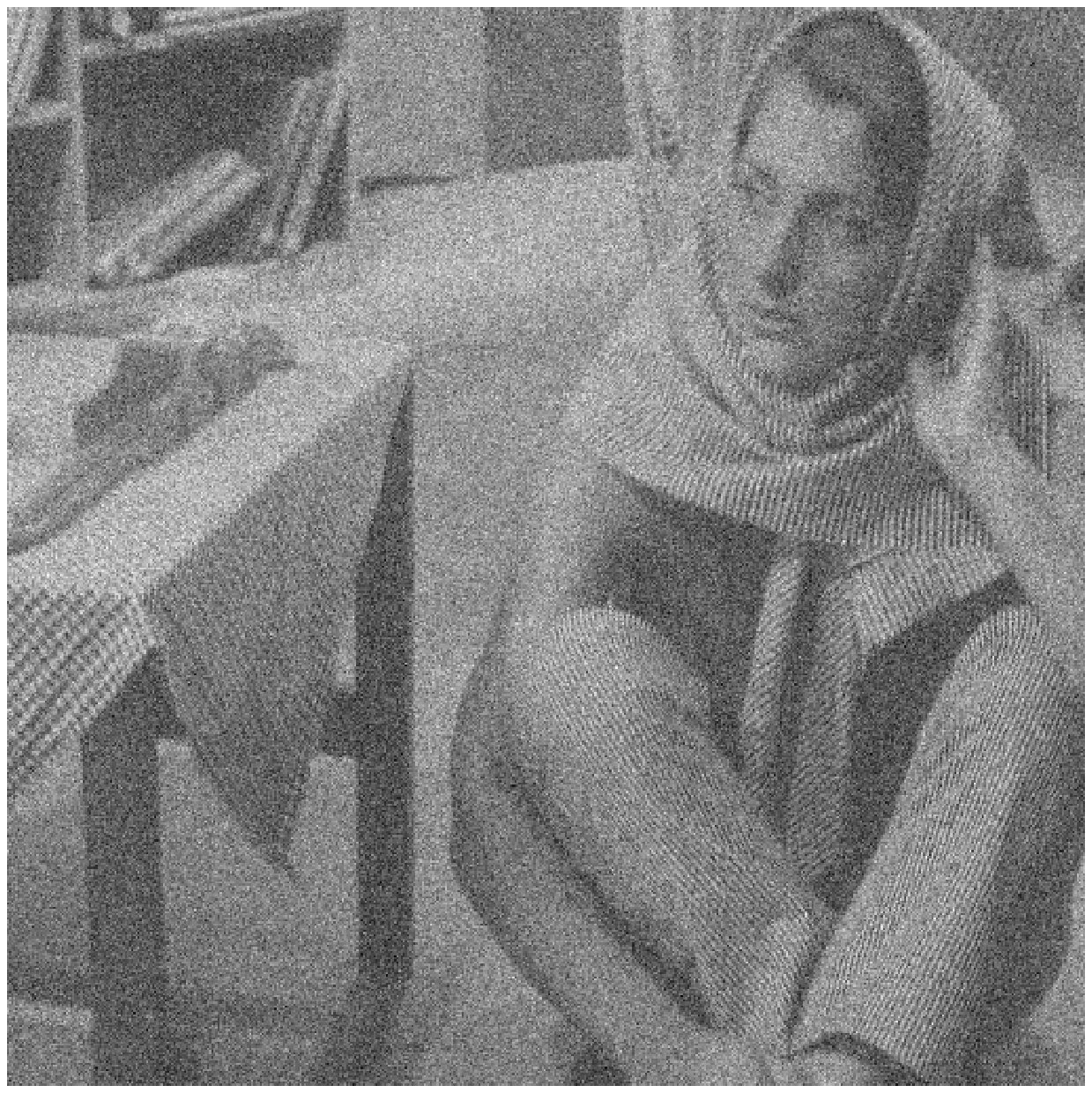}} \\
    \subfloat[\label{nlm} NLM output.]{\includegraphics[width=0.42\linewidth]{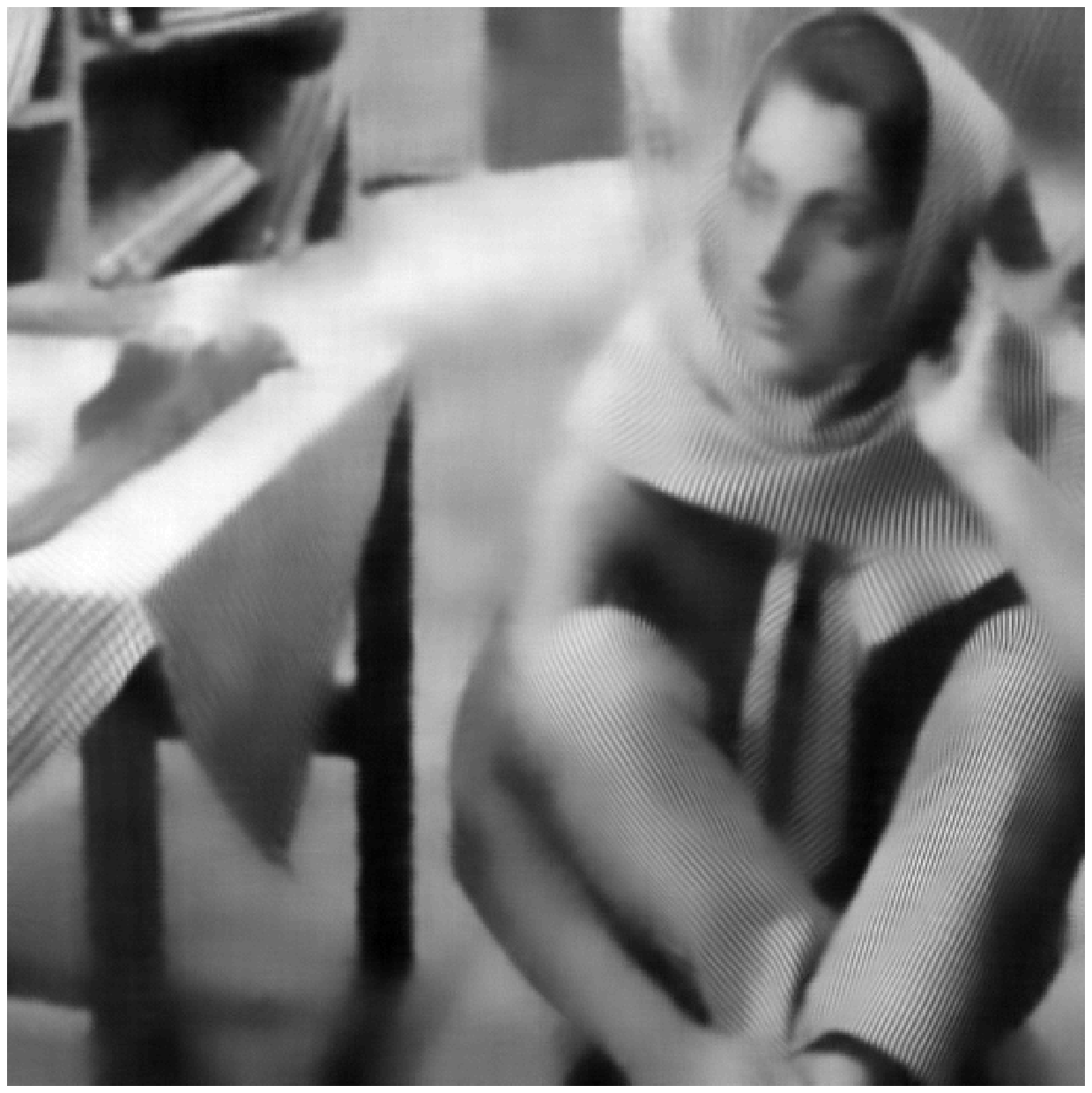}} 
      \subfloat[\label{nlpr} NLPR output ($p=0.1$).]{\includegraphics[width=0.42\linewidth]{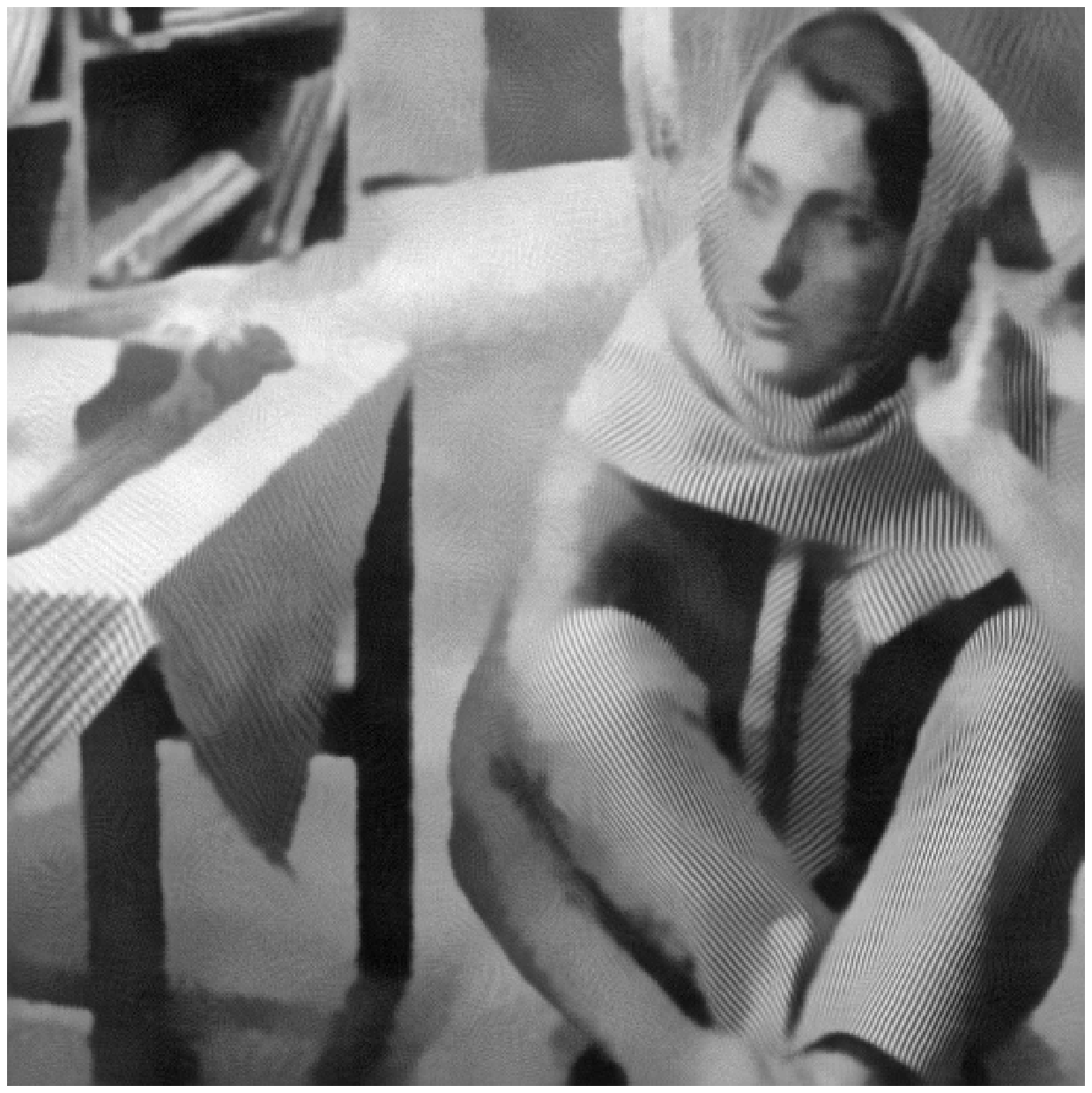}} 
  \caption{Denoising results on the $256 \times 256$ \textit{Barbara} image obtained using NLM and NLPR. The PSNRs are respectively: (b) 16.11 dB, (c) 23.53 dB, and (d) 25.39 dB. Notice that the 
  edges and the texture patterns (on the scarf, pants, and table cloth) are much better restored in NLPR.}
  \label{BarbaraResults}
\end{figure}

\section{Discussion}

\begin{table}
\scriptsize
\caption{Comparison of NLM and NLPR ($p=0.1$) at noise levels $\sigma = 10, 20, \ldots, 100$ (results averaged over $10$ noise realizations)} 
\centering 
\begin{tabular}{l  c rrrrrrrrrr}  

\hline 

\bf{Image} & \bf{Method} &\multicolumn{10}{c}{\bf{PSNR (dB)}} \\

\hline
&NLM    &\bf{34.25}  &29.76        &26.88       &25.21        &24.08       &23.34      &22.81        &22.42      &22.05   &21.80 \\[-1ex]
\raisebox{1.5ex}{\textit{House}} 
&NLPR   &33.23       &\bf{30.23}  &\bf{27.86}   &\bf{26.40}  &\bf{25.45}  &\bf{24.69}  &\bf{24.10}  &\bf{23.52}  &\bf{22.93}  &\bf{22.41}  \\

\hline

&NLM    &\bf{32.38}   &27.38        &24.94   &23.53       &22.65       &22.03      &21.62       &21.30      &21.07  &20.87    \\[-1ex]
\raisebox{1.5ex}{\textit{Barbara}} 
&NLPR   &31.50       &\bf{28.42}    &\bf{26.51} &\bf{25.39}  &\bf{24.57}  &\bf{23.84}  &\bf{23.21}  &\bf{22.60} &\bf{22.06}  &\bf{21.56}  \\

\hline

&NLM      &\bf{30.78}   &26.71   &24.73   &23.64  &22.95  &22.48  &22.12  &21.88  &21.65  &21.45 \\[-1ex]
\raisebox{1.5ex}{\textit{Boat}} 
&NLPR     &30.54  &\bf{27.23}  &\bf{25.50}  &\bf{24.50}  &\bf{23.87}  &\bf{23.40}  & \bf{22.95}  &\bf{22.54} &\bf{22.11}  &\bf{21.68} \\

\hline                          

&NLM      &\bf{31.39}   &\bf{27.90}   &24.78  &22.93  &21.89  &21.14  &20.62  &20.20  &19.88  &19.61  \\[-1ex]
\raisebox{1.5ex}{\textit{Cameraman}} 
&NLPR     &31.17  &27.46  &\bf{25.15}  &\bf{25.15}  &\bf{22.68}  &\bf{22.12}  & \bf{21.67}  &\bf{21.36} &\bf{20.97} &\bf{20.63}  \\

\hline 

&NLM      &\bf{32.34}   &27.66   &24.95  &23.13  &21.89  &21.01  &20.43  &19.98  &19.63  &19.40 \\[-1ex]
\raisebox{1.5ex}{\textit{Peppers}} 
&NLPR     &31.20  &\bf{27.67}  &\bf{25.56}  &\bf{24.18}  &\bf{23.03}  &\bf{22.15}  & \bf{21.62}  &\bf{21.13} &\bf{20.70} &\bf{20.34}  \\

\hline
\end{tabular}
\label{table1}
\end{table}

We compare the PSNRs obtained using NLPR ($p=0.1$) with that of NLM for some standard natural images in Table \ref{table1}. We notice that, for each of the images, NLPR consistently outperforms NLM at large noise levels.
The gain in PSNR is often as large as $2$ dB. The results obtained for \textit{Barbara} using NLM and NLPR are compared in Figure \ref{BarbaraResults}. Note that, as expected, robust regression provides a much 
better restoration of the sharp edges in the image than NLM.  What is probably surprising is that the restoration is superior even in the textured regions. Note, however, that NLM tends to perform better in the smooth regions. For example, we some more noise grains in the smooth regions in Figure \ref{nlpr} compared that
in Figure \ref{nlm}. This suggests that an `adaptive' optimization framework, which combines $\ell_2$ regression (in smooth regions) and $\ell_{(p \leq 1)}$ regression (in the vicinity of edges), might possibly perform better 
than a fixed $\ell_p$ regression. Some other possible extensions of the present work are as follows:  (i) Local convergence analysis of the present IRLS algorithm, and ways of 
improving it; (ii) Possibility of using more efficient numerical algorithms for solving \eqref{Lp_regression}; (iii) Finding better ways of estimating the denoised pixel $\hat{u}_i$ from the estimated patch $\hat{\P}_i$ (the 
projection method used here is probably the simplest); (iv) Use of `better' weights than the ones used in standard NLM \cite{T2009,deVK2011}; and (v) Formulation of a `joint' optimization framework 
for \eqref{Lp_regression}, where the optimization is performed with respect to $w_{ij}$ and $\P$ \cite{Peyre2011}.

\end{document}